\documentclass[runningheads]{llncs}
\usepackage[T1]{fontenc}
\usepackage{multirow}
\usepackage{adjustbox}
\usepackage{tikz}
\usetikzlibrary{shapes.geometric, arrows.meta, positioning, fit, backgrounds}
\usepackage{longtable}
\usepackage{tabularx}
\usepackage{hyperref}

\usepackage{graphicx}
\usepackage{pifont}
\usepackage{newunicodechar}
\usepackage{xcolor}
\usepackage{array}
\usepackage{tcolorbox}

\usepackage{geometry}

\definecolor{promptbg}{RGB}{248,249,250}
\definecolor{promptborder}{RGB}{209,213,219}

\newunicodechar{❌}{\textcolor{red}{\ding{55}}}
\begin{document}
\title{Reasoning with RAGged events: RAG-Enhanced Event Knowledge Base Construction and reasoning with proof-assistants}

\title{Reasoning with RAGged events: RAG-Enhanced Event Knowledge Base Construction and reasoning with proof-assistants}
\author{Stergios Chatzikyriakidis\orcidID{0000-0001-8016-9743} }
\authorrunning{S. Chatzikyriakidis}
\institute{Computational Linguistics and Language Technology lab\\Department of Philology\\ University of Crete\\
\email{stergios.chatzikyriakidis@uoc.gr}}
%\author{Stergios Chatzikyriakidis\orcidID{0000-0001-8016-9743} }
%
%\authorrunning{S. Chatzikyriakidis}
%\institute{Computational Linguistics and Language Technology lab\\Department of Philology\\ University of Crete\\
%\email{stergios.chatzikyriakidis@uoc.gr}}
%
\maketitle              % typeset the header of the contribution
\begin{abstract}
Extracting structured representations of historical events from narrative sources still remains challenging when one constructs them manually. While RDF/OWL reasoners support graph-based reasoning, their expressiveness is limited to restricted fragments of first-order logic. We develop automated models for historical event extraction using large language models (GPT‑4, Claude, Llama 3.2) with three strategies: direct generation, knowledge-graph augmentation, and retrieval-augmented generation (RAG).
Using the 10 first chapters of Thucydides works  as a case study,  we find that different enhancement strategies optimize different performance dimensions rather than providing across the board universal improvements. Direct generation favors coverage, while RAG improves precision but reduces breadth. Model architecture influences this trade-off: large models show stable baselines with incremental RAG benefits, while Llama 3.2 exhibits extreme variance from competitive to catastrophic performance.
To address RDF's expressivity limitations, we develop a translation pipeline converting RDF outputs to Coq proof assistant specifications, enabling temporal arithmetic with BCE dates, multi-step causal inference, and formal validation of domain-specific event types. This demonstrates that optimal enhancement strategies depend on specific application requirements, while establishing foundations for computational humanities combining NLP scalability with formal verification.\footnote{Code and experimental data are available here: \url{https://drive.google.com/drive/folders/1ZkZc3CnNGAdXLUzKly4-FGXiVqk9BpaW?usp=sharing}
.}
%at \url{https://github.com/StergiosCha/RAGGed_events}.
\end{abstract}
\section{Introduction}
Ontologies play a crucial role in computational linguistics and artificial intelligence by codifying knowledge in a structured manner that enables sophisticated information retrieval and reasoning. These formal knowledge representations allow machines to understand semantic relationships between entities, facilitating complex queries that go beyond simple keyword matching. The DBpedia ontology \cite{lehmann2015dbpedia}, with its 768 classes arranged in a subsumption hierarchy and described by over 3,000 properties, exemplifies the rich potential of such structured knowledge repositories for enhancing natural language understanding.

Despite their utility, structured knowledge extraction faces two fundamental challenges that have motivated extensive research in Retrieval-Augmented Generation (RAG). First, manual construction of event knowledge bases remains computationally expensive and labor-intensive. Second, existing RDF/OWL reasoning mechanisms are fundamentally limited to fragments of first-order logic, constraining their ability to handle complex temporal semantics, multi-step causal relationships, and sophisticated analytical queries essential for historical analysis.

While RAG has demonstrated success across various natural language processing tasks, its effectiveness for domain-specific knowledge extraction remains an open empirical question. The prevailing assumption that external knowledge enhancement universally improves performance may not hold when large language models already possess substantial internal knowledge about well-documented domains. Furthermore, the interaction between model architecture and enhancement strategies—particularly for specialized tasks like historical event extraction has received limited systematic investigation.

This paper presents a comprehensive empirical analysis that challenges conventional assumptions about RAG benefits through systematic comparison of automatic historical event knowledge base extraction across multiple LLM architectures and enhancement strategies. We investigate three distinct approaches, base generation without external knowledge, knowledge graph enhancement, and full RAG enhancement, using three state-of-the-art models: GPT-4o, Claude-3.5, and Llama 3.2. Our analysis reveals counterintuitive findings that fundamentally reshape understanding of when and how external knowledge augmentation should be applied.
Our investigation uncovers an "inverse calibration principle" where enhancement effectiveness inversely correlates with model capability: stronger models like GPT-4o and Claude-3.5 achieve superior performance through pure base generation (extracting 36 and 39 comprehensive events respectively), while external enhancement often proves detrimental—introducing hallucinations, reducing coverage, and corrupting semantic accuracy. Conversely, weaker models like Llama 3.2 require external scaffolding but exhibit extreme sensitivity to implementation quality, ranging from competitive performance to catastrophic failure based on enhancement complexity.

These findings reveal that enhancement strategies optimize different performance dimensions rather than providing universal improvements. Base generation excels in comprehensive historical coverage and narrative richness, while RAG enhancement trades breadth for technical precision-improving coordinate accuracy and metadata completeness at the cost of event volume and semantic coherence. This trade-off suggests that optimal strategy selection must be driven by specific use-case requirements rather than blanket enhancement application.

To address the fundamental expressivity limitations of RDF/OWL systems revealed through our extraction analysis, we develop an automated translation pipeline that converts extracted knowledge representations into formal specifications for the Coq proof assistant. This translation enables higher-order reasoning capabilities impossible within traditional semantic web frameworks, including verification of multi-step causal relationships, complex temporal arithmetic with historical dates, and formal proofs about historical causation. The Coq formalization also validates a crucial theoretical finding: RAG-discovered event types that are missing when evaluated against general-purpose knowledge graphs actually represent legitimate domain-specific semantic structures amenable to rigorous mathematical proof.

Our methodology employs historical texts from Thucydides' History of the Peloponnesian War as a controlled domain for systematic evaluation. More specifically, we use the ten first chapters of Thucydides in the English translation by Richard Crawley This choice enables assessment of enhancement strategies against well-documented historical events while revealing how model architecture and domain characteristics interact to determine optimal knowledge integration approaches. The integration of automatic extraction with formal verification establishes a foundation for computational humanities applications that combine modern NLP scalability with mathematical rigor.
The central contribution of this work is demonstrating that enhancement strategy effectiveness depends fundamentally on the interaction between model capability, domain characteristics, and intended application requirements, challenging the field's assumption that more comprehensive external knowledge necessarily leads to better performance. 
\section{Related work}
Retrieval-Augmented Generation (RAG) is an approach that combines the strengths of retrieval-based systems, which fetch relevant information from external sources, and generative models, which synthesize this extracted information into coherent and contextually aware outputs \cite{lewis2020retrieval}. RAG enables large language models (LLMs) to dynamically access up-to-date and domain-specific knowledge, thus improving their factual accuracy, reducing hallucinations, and supporting more nuanced reasoning \cite{gao2023retrieval}.

RAG models have achieved state-of-the-art results across various tasks, including open-domain question answering, information retrieval, and knowledge-grounded text generation \cite{karpukhin2020dense}. By leveraging external memory (such as Wikipedia or specialized knowledge bases), RAG architectures outperform purely parametric models, particularly in generating more specific, diverse, and factually grounded responses \cite{izacard2021leveraging}. The typical RAG workflow involves retrieving relevant documents or passages, fusing this information with the input prompt, and then generating the final output using an LLM \cite{lewis2020retrieval}.

Beyond general question answering, RAG techniques have been successfully applied to information extraction (IE) tasks, including named entity recognition, relation extraction, and event extraction \cite{yu2022retrieval}. In these settings, RAG enhances both performance and interpretability by grounding extractions in retrieved, contextually relevant evidence. For instance, in named entity recognition, retrieval of similar sentences improves semantic representations, while in event argument extraction, retrieval-augmented models help capture cross-argument dependencies and improve few-shot learning performance \cite{li2023}.

Recent studies have specifically explored RAG for event argument extraction, a task that involves identifying and classifying the participants and attributes of events described in text \cite{du2020event}. Retrieval-augmented generative models, such as R-GQA \cite{du2022} and DRAGEAE \cite{he2024demonstration}, retrieve similar QA pairs or demonstration examples to inform the generation process, resulting in substantial improvements over extractive and standard generative baselines. These approaches are particularly effective in scenarios with limited annotated data, as they can leverage demonstrations from related contexts to guide extraction \cite{ma2023few}.

RAG has been used in temporal event relation extraction  to address the inherent ambiguity of temporal semantics in text \cite{zhou2024event}. By retrieving relevant knowledge from LLMs and integrating it into prompt templates, RAG-based methods achieve superior performance in extracting and reasoning about event temporal relations \cite{liu2023rag}.

Another relevant line of work is  grounding RAG   in structured ontologies, for example domain-specific knowledge graphs \cite{baek2023knowledge}. Ontology-Grounded RAG (OG-RAG) approaches basically construct hypergraph representations of domain documents and cluster factual knowledge according to ontological relationships \cite{christmann2023ontology,luo2025hypergraphrag}. 

Recent comparative studies of RAG configurations reveal that the relationship between retrieval complexity and performance is non-linear \cite{cuconasu2024power}. Our work contributes to this emerging understanding by demonstrating that for domain-specific knowledge extraction, particularly in well-documented historical domains, moderate retrieval strategies can outperform both minimal and maximal external knowledge integration \cite{ovadia2023fine}. This challenges the prevalent assumption in the RAG literature that more comprehensive retrieval necessarily leads to better performance, highlighting the importance of domain-specific optimization in RAG system design \cite{asai2023retrieval}.

Lastly, Coq has been used to formalize natural language temporal and aspectual semantics by \cite{bernardy2021} as part of a symbolic Natural Language Inference system, achieving state-of-the-art results for symbolic approaches. In general, Coq has been shown to be a powerful tool to formalize natural language semantics \cite{chatzikyriakidis2016}.

Our proposal has similarities with some of the related work, but it is unique in its main approach. What we offer is twofold: a) the use of RAG for automatic event knowledge base extraction with systematic comparison of different RAG and non-RAG configurations, and b) a translation of the resulting extracted events into the Coq proof assistant, enabling advanced temporal and causal reasoning beyond what is possible with simple RDF languages \cite{hogan2021knowledge,paulheim2017knowledge}.

Rather than contradicting the general insight that accurate, targeted knowledge improves LLM performance, our experiments refine it: we show that injection of excessive or loosely filtered context can degrade extraction quality, especially in already knowledge-rich models. We call this the \emph{inverse calibration principle}.

\section{Methodology}
The proposed approach implements a two-phase pipeline for transforming unstructured historical narratives into formal knowledge representations. Phase 1 employs a number of NLP techniques for semantic event extraction, while Phase 2 performs logic translation to enable automated reasoning and formal verification:

\begin{itemize}
\item \textbf{Phase 1}
\begin{itemize}

   \item Event boundary detection and segmentation
   \item Agent identification with semantic role assignment
   \item Geographical entity resolution and coordinate mapping
   \item Temporal expression normalization
   \item Outcome extraction and causal relationship identification
   \item RDF knowledge graph construction
\end{itemize}\end{itemize}

\begin{itemize}
\item \textbf{Phase 2}
\begin{itemize}

   \item RDF-to-Coq inductive type conversion
   \item Higher-order temporal logic implementation
   \item Causal inference framework integration
   \item Proof-assistant compatibility for formal verification
\end{itemize}\end{itemize}

In phase 1, we  use three enhancement strategies across three LLM architectures using historical text from Thucydides' \textit{History of the Peloponnesian War} (first 9 chapters). Our knowledge sources include the DBpedia ontology  that involves 768 classes and 3,000+ properties, and the specialized LACRIMALit ontology) for Ancient Greek geographical locations \cite{papadopoulou2022lacrimalit}. External knowledge retrieval includes the Wikidata and the DBpedia SPARQL endpoints, as well as the ConceptNet API. The three models we use are GPT-4o (large commercial model), Claude-3.5 Sonnet (alternative large-scale architecture), and Llama 3.2 (3.2B parameter model via local Ollama deployment). All models use deterministic settings (temperature=0) for consistency and use the same chunking parameters. We use three variations on these models, base-generation, generation with knowledge enhancement, and generation with knowledge enhancement and RAG. 

\textbf{True Base Generation (Control):} Pure LLM inference without external knowledge augmentation. For this approach, the input text is segmented into chunks, and the model processes these chunks one by one, relying solely on its internal knowledge to complete the extraction. All APIs, knowledge graphs, location enrichment, and RAG functionality are explicitly disabled for baseline measurement. However, the baseline uses the structured prompt shown in figure \ref{fig:system_prompt}.
\begin{figure}[h!]
\centering
\begin{tcolorbox}[
    colback=promptbg,
    colframe=promptborder,
    boxrule=1pt,
    arc=3pt,
    left=10pt,
    right=10pt,
    top=10pt,
    bottom=10pt,
    title={\textbf{System Prompt for Historical Event Extraction}},
    fonttitle=\large\bfseries,
    coltitle=black,
    colbacktitle=white
]
\small
\textbf{You are extracting historical events from text using ONLY the information provided in the text chunk.}

\vspace{0.3cm}
\textbf{TEXT CHUNK \{chunk\_num\} TO ANALYZE:}\\
\textit{\{chunk\}}

\vspace{0.3cm}
\textbf{ENTITIES FOUND:} \{entities\}\\
\textbf{LOCATIONS FOUND:} \{locations\}

\vspace{0.3cm}
\textbf{TASK:} Extract historical events using text information and making REASONABLE INFERENCES.

\vspace{0.3cm}
\textbf{REQUIREMENTS:}
\begin{enumerate}
\item Extract ONLY events explicitly mentioned in the text chunk
\item Use information directly stated in the text
\item MAKE REASONABLE INFERENCES from context clues
\item Infer coordinates, countries, regions from textual context
\end{enumerate}

\vspace{0.3cm}
\textbf{OUTPUT FORMAT:}\\
\texttt{ste:Event\{chunk\_num\}\_1 a ste:Event ;}\\
\texttt{~~~~ste:hasType "event description" ;}\\
\texttt{~~~~ste:hasAgent "person/group" ;}\\
\texttt{~~~~ste:hasTime "time period" ;}\\
\texttt{~~~~ste:hasLocation "location name" ;}\\
\texttt{~~~~ste:hasLatitude "37.9838" ;}\\
\texttt{~~~~ste:hasLongitude "23.7275" ;}\\
\texttt{~~~~ste:hasCountry "Greece" ;}\\
\texttt{~~~~ste:hasRegion "Attica" ;}\\
\texttt{~~~~ste:hasLocationSource "inferred" ;}\\
\texttt{~~~~ste:hasResult "outcome" .}

\end{tcolorbox}
\caption{LLM System Prompt for Historical Event Extraction}
\label{fig:system_prompt}
\end{figure}

\textbf{Knowledge Enhanced (No-RAG):} Here, we use knowledge integration but no RAG. We attempt entity extraction, knowledge graph retrieval queries to Wikidata, DBpedia, and ConceptNet. Lastly,  we attempt location enrichment that follows the order LACRIMALit ontology → Wikidata → DBpedia, stopping upon successful coordinate retrieval.

The placeholders \{entities\} and \{locations\} are automatically populated 
by a preprocessing pipeline: named entities are extracted via regex/NER heuristics 
(limit 25 per chunk) and candidate locations are identified through a LocationExtractor 
that consults regex patterns, the LACRIMALit ontology, and then Wikidata/DBpedia. 
RAG chunks are obtained via FAISS with MiniLM embeddings; we used chunk size 
1,500–2,000 tokens with 200 character overlap. 
Evaluation combined manual expert assessment of correctness with quantitative 
counts of extracted events, coordinate precision (Geo distance from known sites), 
and RDF formatting completeness.

\begin{figure}[h!]
\centering
\begin{tikzpicture}[
    node distance=2cm,
    box/.style={rectangle, rounded corners, fill=blue!20, draw, minimum width=2.5cm, minimum height=1cm, text centered},
    arrow/.style={->, >=stealth, thick}
]

% Pipeline nodes
\node[box] (input) {Input Text};
\node[box, right=of input] (chunk) {Text Chunking};
\node[box, right=of chunk] (kg) {Knowledge Graph Retrieval};
\node[box, below=of kg] (rag) {RAG Context Assembly};
\node[box, left=of rag] (llm) {Enhanced LLM Prompt};
\node[box, left=of llm] (output) {RDF Output};

% Arrows
\draw[arrow] (input) -- (chunk);
\draw[arrow] (chunk) -- (kg);
\draw[arrow] (kg) -- (rag);
\draw[arrow] (rag) -- (llm);
\draw[arrow] (llm) -- (output);

\end{tikzpicture}
\caption{The RAG Pipeline}
\label{fig:rag}
\end{figure}
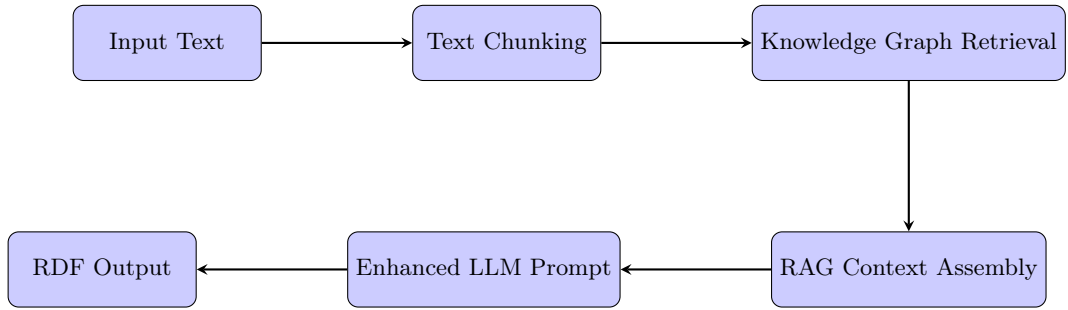

\textbf{RAG Enhancement:} This adds a Retrieval-augmented generation layer that combines vector similarity with knowledge graph enhancement. Implementation uses FAISS vector store with HuggingFace paraphrase-multilingual-MiniLM-L12-v2 embeddings, and different parameters for k. The RAG pipeline is shown in figure \ref{fig:rag}.

\subsection{Implementation and Evaluation}
Our system employs cascading queries (local ontology → Wikidata → DBpedia) with MD5-based caching and rate limiting to minimize API calls. Evaluation combines quantitative metrics (event coverage, coordinate precision, RDF completeness) with qualitative assessment of semantic depth and ontological classification. All outputs follow standardized RDF/Turtle format using consistent properties: \texttt{ste:hasType}, \texttt{ste:hasAgent}, \texttt{ste:hasTime}, \texttt{ste:hasLocation}, \texttt{ste:hasLatitude/Longitude}, and \texttt{ste:hasResult}. \texttt{ste:hasType}, \texttt{ste:hasAgent}, \texttt{ste:hasTime}, \texttt{ste:hasLocation}, \texttt{ste:hasLatitude/Longitude}, and \texttt{ste:hasResult} to enable direct comparison across enhancement strategies.

\subsection{Enhancement Strategy Comparison}

Our systematic investigation examined how different knowledge integration approaches affect extraction performance across model architectures. We evaluated the three enhancement strategies using identical prompting frameworks to ensure fair comparison. The results are shown in Table \ref{results}. Our analysis reveals distinct performance trade-offs that challenge conventional assumptions about external knowledge benefits. The results demonstrate that enhancement strategies are detrimental with stronger models and helpful with less powerful models. Base generation consistently achieved maximum event coverage (GPT-4: 36 events, Claude: 39 events) with rich narrative detail and comprehensive historical scope, producing almost clean RDFs in terms of syntax (the only thing missing is xsd format, that was missing in the prompt for base but included in the other architectures). Knowledge-base and RAG consistently produced fewer events, while at the same time introduced hallucinations into the text. For example, the coordinates for the battle of Sybota are correctly attributed to Sybota in base GPT-4o, but incorrectly given in Knowledge Enhancement GPT-4o. A bit of research reveals that the coordinates used belong to Ambracia, one of the cities that participated in the battle of Sybota. Claude-RAG conflates three distinct temporal layers: ancient events (431 BC battles, sieges), authorship (Thucydides writing about the war), and modern publishing (Crawley translations, Project Gutenberg updates), extracting all as if they were contemporary Peloponnesian War events. Additionally includes metadata like \textit{hasLLM: Claude4} as RDF properties.

LLAMA exhibits the most dramatic variation in performance, revealing a brittle failure mode under complex enhancement conditions. Base LLAMA produces only 10 events with poor formatting, establishing a low baseline. Simple RAG configurations (RAG 1-2) deliver substantial improvements, boosting output to 37-38 events with better structure, demonstrating that knowledge augmentation can effectively compensate for weaker model capabilities. However, this improvement exhibits catastrophic failure at higher complexity levels: more sophisticated RAG setups cause complete breakdown, with RAG 3 producing only 6 events and RAG 4 achieving zero correct event structures. This suggests that weaker models benefit from controlled knowledge injection but lack the architectural robustness to handle complex multi-source reasoning without systematic collapse.

These findings reveal a nuanced inverse calibration principle that operates along two critical dimensions: enhancement intensity and model capability. Recent empirical work has demonstrated that stronger retrievers paradoxically cause more performance degradation, with performance exhibiting an "inverted-U pattern" where initial improvements are followed by sharp decline \cite{jin2024long}. Our results extend this phenomenon beyond retrieval to encompass broader knowledge integration strategies, revealing that the relationship between model strength and enhancement benefit follows a complex inverse calibration where stronger models perform optimally with minimal augmentation while weaker models require carefully bounded enhancement.

The LLAMA trajectory, i.e. from baseline failure (10 events) through optimal performance (37-38 events) to catastrophic collapse (0 events), illustrates a fragility threshold in knowledge integration. This aligns with findings that weaker models are more susceptible to ``hard negatives'' and can only handle effective knowledge augmentation up to specific complexity limits before experiencing systematic breakdown \cite{jin2024long}. The  improvement from base to simple RAG demonstrates that weaker architectures can benefit substantially from controlled knowledge injection, effectively compensating for  knowledge deficits. However, the catastrophic failure at higher RAG complexity levels reveals that these same models lack the architectural robustness to handle multi-source reasoning without semantic collapse.

Conversely, stronger models like GPT-4o and Claude exhibit the opposite pattern: their sophisticated parametric knowledge and reasoning capabilities are disrupted rather than enhanced by external knowledge sources. This supports research showing that stronger models exhibit different relationships between internal knowledge and retrieved context, often prioritizing external information even when it contradicts their more accurate training data \cite{wu2024}. The coordinate misattribution in GPT-4o and temporal conflation in Claude-RAG exemplify how knowledge augmentation can introduce noise that overwhelms these models'  pattern recognition capabilities, causing them to move away from their reliable internal representations in favor of potentially corrupted external sources. This phenomenon reflects the theoretical framework where knowledge integration increases cognitive load and introduces semantic drift, particularly problematic for models with sophisticated internal knowledge structures that can be destabilized by conflicting external information \cite{farquhar2024}.

\begin{table}[h!]
\centering
\scriptsize
\begin{tabularx}{\textwidth}{|l|l|c|c|c|X|}
\hline
\textbf{Model} & \textbf{Configuration} & \textbf{Events} & \textbf{Precision} & \textbf{RDF} & \textbf{Issues} \\
\hline
GPT & Base Generation & 36 & Excellent & Excellent & Missing xsd (due to prompt) \\
\hline
GPT & No RAG + KG & 30 & Good (some Hallucinations) & Excellent & Some empty coordinates \\
\hline
GPT & RAG + Multi KG 1 & 23 & Good (some Hallucinations) & Excellent & Some empty coordinates\\
\hline
GPT & RAG + Multi KG 2 & 23 & Good (some Hallucinations)  & Excellent & some empty coordinates \\
\hline
GPT & RAG + Multi KG 3 & 20 & Good (some Hallucinations) & Perfect & No syntax issues \\
\hline
GPT & RAG + Multi KG 4 & 26 & Good (some Hallucinations)  & Perfect & No syntax issues \\
\hline
\hline
Claude 4 & Base Generation & 39 & Excellent & Excellent & Missing xsd (due to prompt) \\
\hline
Claude 4 & No RAG + KG & 20 & Fair (hallucinations) & Good  & Repeats event ids \\
\hline
Claude 4 & RAG + Multi KG 1 & 10 & Poor (Many Hallucinations) & Good & Missing coordinate fields \\
\hline
Claude 4 & RAG + Multi KG 2 & 16 & Poor (Many Hallucinations) & Good & Missing coordinate fields \\
\hline
Claude 4 & RAG + Multi KG 3 & 19 & Poor (Many Hallucinations)  & Good & Missing coordinate fields\\
\hline
Claude 4 & RAG + Multi KG 4 & 19 & Poor (Many Hallucinations)  & Good & Missing coordinate fields \\
\hline
\hline
LLAMA & Base Generation & 10 & Fair & Poor & Many formatting errors \\
\hline
LLAMA & No RAG + KG & 19 & Good & Fair & Some formatting errors \\
\hline
LLAMA & RAG + Multi KG 1 & 38 & Good & Good & Some missing coordinates and repeated event ids\\
\hline
LLAMA & RAG + Multi KG 2 & 37 & Good & Good & Some missing coordinates and repeated event ids \\
\hline
LLAMA & RAG + Multi KG 3 & 6 & Poor & Poor & Many syntax errors \\
\hline
LLAMA & RAG + Multi KG 4 & 0 & Bad & Bad & No correct event structure \\
\hline
\end{tabularx}
\caption{Model Performance Comparison. Here “No RAG + KG” refers to enrichment from a single source (primary triple retrieval),
while “RAG + Multi KG” combines context from all three sources 
(Wikidata, DBpedia, ConceptNet) with vector similarity retrieval.}
\label{results}
\end{table}

Beyond performance metrics, RAG in some cases also appears to corrupt the semantic accuracy of extracted events and disrupt ontological consistency when evaluated against standard knowledge graphs. While some extracted events align with DBpedia subtypes (e.g., SocietalEvent), most domain-specific event descriptions fall outside DBpedia's taxonomy (e.g., there is no WarEvent or MigrationEvent in the standard ontology). The prompt's reference to \textit{dbp:SpecificEventType} reveals this ontological gap, i.e. models frequently default to this generic classification when encountering historically meaningful events that require finer semantic distinctions. However, these apparent inconsistencies reflect a fundamental granularity mismatch rather than genuine semantic errors. RAG consistently discovers domain-specific event types and relationships that exceed general-purpose ontologies' coverage. While these extractions violate DBpedia's constraints, they often capture historically accurate and semantically coherent distinctions essential for domain analysis. This subevent discovery would be helpful when we will discuss our translation procedure to Coq.

Lastly, other  cases where RAG messes up is cases where a broader historical context from the text is taken into consideration but the information presented is conflated and inaccurate like you see in the example below:

\begin{table}[htbp]
\centering
\caption{RDF Event Extraction Example: Temporal Conflation in Claude-RAG}
\begin{tabular}{|l|l|}
\hline
\textbf{Property} & \textbf{Value} \\
\hline
\texttt{rdf:type} & \texttt{ste:Event, dbp:SpecificEventType} \\
\hline
\texttt{ste:hasType} & The outbreak of the Peloponnesian War \\
\hline
\texttt{ste:hasAgent} & Thucydides \\
\hline
\texttt{ste:hasTime} & 431 BC \\
\hline
\texttt{ste:hasLocation} & Greece \\
\hline
\texttt{ste:hasLatitude} & 39.0742 (\texttt{xsd:double}) \\
\hline
\texttt{ste:hasLongitude} & 21.8243 (\texttt{xsd:double}) \\
\hline
\texttt{ste:hasCountry} & Greece \\
\hline
\texttt{ste:hasRegion} & Hellas \\
\hline
\texttt{ste:hasLocationSource} & wikidata \\
\hline
\texttt{ste:hasResult} & The greatest movement yet known in history, \\
& involving the Hellenes and a large part of \\
& the barbarian world \\
\hline
\texttt{ste:hasRAGContext} & yes \\
\hline
\end{tabular}
\label{tab:rdf-event-example}
\end{table}

The example in table \ref{tab:rdf-event-example} illustrates not only  the temporal conflation problem where Thucydides (the historian who documented the war) is incorrectly identified as the agent causing the Peloponnesian War itself, but also incorrectly identifies the war as \textit{the greatest movement in history involving the Hellenes and large part of the barbarian world}. This clearly demonstrates how RAG enhancement can fundamentally corrupt the semantic relationships between historical actors and events instead of make them more precise and enriching them. 

\section{Beyond RDF: Enabling Higher-Order Reasoning}

While our event extraction system successfully produces structured RDF knowledge graphs from historical texts, these representations encounter fundamental computational barriers when subjected to more fine-grained demands. Our RAG evaluation demonstrates that no matter the enhancement strategy, be it base generation, knowledge augmentation, or full RAG, the RDF knowledge graphs we get   identical limitations when processed by conventional reasoning systems. More fine-grained temporal dependencies, multi-layered causal networks, and emergent narrative patterns common in historical discourse require reasoning architectures that can extend beyond systems, albeit extremely useful ones, that are limited to fragments of first-order logic.

\subsection{Computational Limitations of RDF/OWL Systems}

Contemporary RDF/OWL reasoning engines operate within restrictive logical fragments that prove systematically inadequate for historical analysis. These systems are constrained to decidable subsets of first-order logic, fundamentally limiting their capacity to express and verify complex historical relationships. Consider the cascading causal sequence linking the Siege of Plataea (429-427 BCE) to Pericles' political decline: the prolonged siege concentrated Athenian populations within city walls, creating conditions that amplified plague transmission, which subsequently shifted political sentiment and undermined Pericles' leadership over multiple years. While RDF can represent these as discrete events with simple causal predicates (\texttt{:causedBy}), it cannot formally express the multi-step temporal dependencies or verify the logical consistency of such cascading relationships.

Traditional SPARQL queries lack the expressive power for inductive reasoning patterns essential to historical analysis. Questions such as ``Do successful military campaigns correlate with subsequent political stability?'' require pattern matching across temporal sequences and statistical inference over event distributions, capabilities that exceed SPARQL's graph pattern matching paradigm. Furthermore, counterfactual reasoning  demands modal logic constructs entirely absent from RDF/OWL specifications.

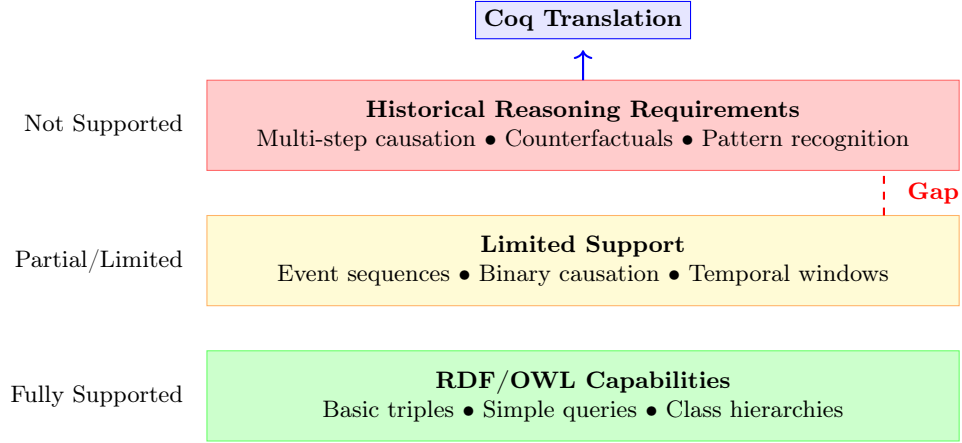
\begin{figure}[htbp]
\centering
\begin{tikzpicture}[
    layer/.style={rectangle, minimum width=10cm, minimum height=1.2cm, align=center},
    supported/.style={layer, fill=green!20, draw=green!60},
    partial/.style={layer, fill=yellow!20, draw=orange!60},
    unsupported/.style={layer, fill=red!20, draw=red!60}
]
% RDF/OWL Supported (Bottom)
\node[supported] (rdf) at (0, 0) {
    \textbf{RDF/OWL Capabilities} \\
    Basic triples $\bullet$ Simple queries $\bullet$ Class hierarchies
};
% Partially Supported (Middle)
\node[partial] (partial) at (0, 1.8) {
    \textbf{Limited Support} \\
    Event sequences $\bullet$ Binary causation $\bullet$ Temporal windows
};
% Historical Requirements (Top)
\node[unsupported] (required) at (0, 3.6) {
    \textbf{Historical Reasoning Requirements} \\
    Multi-step causation $\bullet$ Counterfactuals $\bullet$ Pattern recognition
};
% Labels
\node[left] at (-5.2, 0) {\small Fully Supported};
\node[left] at (-5.2, 1.8) {\small Partial/Limited};
\node[left] at (-5.2, 3.6) {\small Not Supported};
% Gap indicator
\draw[thick, dashed, red] (4, 2.4) -- (4, 3.0);
\node[right, red] at (4.2, 2.7) {\textbf{Gap}};
% Solution
\node[draw=blue, fill=blue!10] at (0, 5) {
    \textbf{Coq Translation}
};
\draw[->, blue, thick] (0, 4.2) -- (0, 4.6);
\end{tikzpicture}
\caption{RDF/OWL capabilities versus historical reasoning requirements. The reasoning gap between limited RDF expressivity and complex historical analysis necessitates translation to formal verification systems.}
\label{fig:rdf-limitations}
\end{figure}

It is true that extensions such as SWRL or DL-safe Horn rules can cover additional 
entailed facts beyond pure RDFS. However, these rule regimes remain essentially first-order, 
lacking facilities for dependent types, integer date arithmetic over BCE conventions, 
or higher-order causal composition, needed for careful and fine-grained reasoning over events. Our use of Coq thus complements rather than 
replaces SWRL-style reasoning: SWRL is well-suited for monotonic rule expansions, 
while Coq enables formal arithmetic and multi-step causal proofs that exceed those capabilities.

% Add this single paragraph at the end of Section 3, after the temporal conflation table example

\subsection{Formal Translation Architecture}

We develop an automated translation pipeline that converts extracted RDF/Turtle representations into formal specifications for the Coq proof assistant. This architecture enables higher-order reasoning while preserving the semantic richness captured during RAG-enhanced extraction, validating our earlier findings about ontological discovery versus hallucination.

The translation system comprises four integrated components: (1) an RDF parsing module that extracts event structures, temporal relationships, and participant roles from Turtle serializations; (2) a type inference engine that maps RAG-discovered event classes to formal Coq inductive types, demonstrating that fine-grained classifications like \texttt{MigrationEvent} and \texttt{WarEvent} represent legitimate semantic structures and extensions of a pre-existing ontology; (3) a temporal logic translator that converts chronological relationships into mathematical constraints using dependent types; and (4) a theorem generation system that produces verifiable propositions about historical causation, enabling formal verification of complex historical hypotheses.

\subsection{Higher-Order Reasoning Capabilities}

The Coq formalization unlocks analytical capabilities impossible within RDF frameworks. Multi-step causal reasoning becomes expressible through function composition and dependent types:

\begin{verbatim}
Definition cascading_causation (e1 e2 e3 : Event) : Prop :=
  temporal_precedes e1 e2 /\ may_have_caused e1 e2 /\
  temporal_precedes e2 e3 /\ may_have_caused e2 e3 /\
  same_historical_context e1 e3.
\end{verbatim}

This enables formal verification of historical propositions that require transitive reasoning across multiple temporal periods. Inductive definitions allow pattern recognition across event sequences, while dependent types capture contextual constraints that ensure logical consistency. The system can formally prove or disprove complex historical claims, transforming historical analysis from interpretive discourse into rigorous logical investigation.

Crucially, this translation validates our ontological discovery thesis: RAG-generated event types that violate DBpedia constraints integrate seamlessly into Coq's rich type system, demonstrating their semantic legitimacy. The apparent ontological inconsistencies represent valuable domain-specific knowledge structures that formal reasoning systems can exploit for sophisticated historical analysis.

\subsection{Translation Process}

The automated translation from RDF/Turtle representations to Coq specifications operates through a four-stage pipeline that systematically transforms extracted event data into formal logical constructs while validating ontological discoveries.

\textbf{Stage 1: Ontological Discovery Validation.} The translation process begins by analyzing the class hierarchy extracted from RDF triples, distinguishing between standard DBpedia classes and RAG-discovered event types. Our validation algorithm assesses each non-standard class using semantic indicators (presence of domain-specific terms like ``War'', ``Migration'', ``Siege'') and contextual coherence (temporal consistency, specific participants, geographical anchoring). This validation step demonstrates that apparent ontological gaps  represent legitimate domain knowledge rather than hallucinations—for instance, \texttt{ET\_MigrationEvent} represents historically accurate fine-grained event classification that exceeds DBpedia's granularity.

\textbf{Stage 2: Type System Construction.} The translator constructs Coq's rich type system by mapping RDF classes to inductive types with formal subclass relationships. RAG-discovered event types integrate seamlessly into this hierarchy, validating our thesis that these represent coherent semantic structures. The system generates dependent types for temporal reasoning, enabling formal arithmetic on BC dates through Coq's Z library:

\begin{verbatim}
Definition TimePoint := (Z * nat * nat)%type.
Definition is_bc_event (e : Event) : bool :=
  match time e with
  | Some (y, _, _) => Z.ltb y 0%Z
  | None => false
  end.
\end{verbatim}

\textbf{Stage 3: Relationship Formalization.} Complex event relationships that exceed RDF's expressive capacity become formally verifiable propositions. Multi-step causal reasoning emerges through function composition and dependent types, enabling verification of cascading historical causation impossible within SPARQL's pattern matching paradigm:

\begin{verbatim}
Definition cascading_causation (e1 e2 e3 : Event) : Prop :=
  event_before e1 e2 = true /\ may_have_caused e1 e2 = true /\
  event_before e2 e3 = true /\ may_have_caused e2 e3 = true /\
  same_historical_context e1 e3 = true.
\end{verbatim}

\textbf{Stage 4: Theorem Generation.} The final stage generates formal theorems that validate both the semantic coherence of RAG discoveries and enable complex historical reasoning. The \texttt{rag\_discoveries\_coherent} theorem formally proves that RAG-extracted events with proper temporal, participant, and class structure represent valid domain knowledge, while another theorem demonstrates how domain-specific classifications enhance causal inference capabilities.

This translation architecture validates our central claim: RAG-discovered event types that do not appear in the event database but can be used to extend the ontology to represent valuable semantic innovations that formal reasoning systems can exploit for sophisticated historical analysis beyond the capabilities of traditional semantic web frameworks.

\subsection{Toward Practical Applications: Foundation for Future Development}

The Coq translation establishes a foundation for applications that would be impossible with traditional RDF/OWL systems, though significant development remains before practical deployment. Our system demonstrates the feasibility of formally representing complex historical relationships and verifying basic logical consistency within extracted event structures. While current applications require Coq expertise, the architecture proves that RAG-discovered ontologies can support formal reasoning about historical causation.

The primary contribution lies in demonstrating that apparent RAG ``hallucinations'' represent legitimate ontological discoveries amenable to formal verification. This reframes evaluation criteria from conformity to existing knowledge graphs toward recognizing domain-specific semantic innovations. Rather than dismissing fine-grained event classifications as errors, our approach validates their semantic coherence through formal mathematical proof.

The translation architecture opens several promising research avenues. Future work could develop domain-specific logical frameworks that capture historical reasoning patterns without requiring full Coq proficiency. Automated consistency checking tools for digital humanities scholars represent another promising direction, as do hybrid systems that combine statistical correlation with formal logical constraints for causal analysis. Additionally, user-friendly interfaces could eventually expose formal verification capabilities to historians without formal logic training.

Several limitations constrain immediate practical deployment. The approach requires substantial formal logic expertise, limiting accessibility for working historians. The system currently handles relatively simple historical relationships, and scaling to complex historical narratives with multiple conflicting sources remains challenging. The translation process also requires manual verification of generated Coq definitions to ensure correctness.

This integration of automatic extraction with formal verification represents a foundational step toward rigorous computational humanities tools. While not immediately deployable for working historians, the methodology demonstrates that modern NLP techniques can be grounded in formal logical systems, suggesting promising directions for future tool development that could eventually provide both scalability and mathematical rigor for historical analysis.

\section{Conclusions, Limitations and Future Work}

Our systematic evaluation of RAG-enhanced historical event extraction reveals that enhancement strategy effectiveness depends fundamentally on model architecture and domain characteristics rather than following universal patterns. Stronger models like GPT-4o and Claude-3.5 achieve optimal performance through base generation (extracting 36 and 39 events respectively), while external enhancement often proves detrimental—introducing hallucinations and reducing coverage. Conversely, weaker models like Llama 3.2 benefit from controlled knowledge augmentation but exhibit catastrophic failure under complex RAG configurations, demonstrating an inverse calibration principle where model capability inversely correlates with enhancement benefit.

These findings challenge the assumption that comprehensive retrieval necessarily improves performance. Enhancement strategies optimize different dimensions: base generation excels in coverage and narrative richness, while RAG trades breadth for technical precision. This suggests that optimal strategy selection must be driven by specific application requirements rather than blanket enhancement approaches.

Beyond extraction optimization, our work reveals that RAG systems discover domain-specific ontological structures that exceed general-purpose knowledge graphs. Fine-grained event types like ET\_MigrationEvent and ET\_WarEvent, which appear as gaps when evaluated against DBpedia, represent legitimate semantic innovations that capture historically accurate distinctions essential for domain analysis.

To address the expressivity limitations of RDF/OWL systems revealed through our extraction analysis, we developed an automated translation pipeline converting extracted representations into Coq proof assistant specifications. This enables higher-order reasoning impossible within traditional semantic web frameworks: multi-step causal verification, temporal arithmetic with BCE dates, and formal validation of RAG-discovered event types. The successful integration of these domain-specific classifications into Coq's type system validates their semantic legitimacy through mathematical proof.

A main limitation of this current work is that  our evaluation focuses on classical Greek historical texts, providing controlled conditions but limiting generalizability. Future work should systematically evaluate these patterns across diverse historical periods, languages, and narrative styles to establish broader applicability of the inverse calibration principle. Additionally, while our Coq translation demonstrates formal reasoning feasibility, practical deployment requires accessible interfaces that reduce the formal logic expertise barrier for working historians.

This work establishes a foundation for computational humanities that combines NLP scalability with formal verification rigor. For knowledge-rich domains like classical history, the internal knowledge  of LLMs is shown to be remarkably comprehensive, suggesting that external enhancement requires careful domain-specific evaluation rather than universal application. Lastly, the integration of automatic extraction with formal reasoning opens promising directions for tools that can support both large-scale processing and rigorous analytical workflows demanded by historical scholarship.
\section*{Acknowledgements}
The author gratefully acknowledges funding from Amazon (project: Neural-Symbolic Integration for Enhanced Natural Language Processing (NIELS)). 

\end{document}